\pdfoutput=1
\documentclass[conference]{ieeeconf}

\IEEEoverridecommandlockouts    
\usepackage[utf8]{inputenc}

\usepackage{xcolor}

\usepackage{array}
\usepackage{graphics}

\usepackage{blindtext, graphicx}

\usepackage{graphicx}
\usepackage{subfigure}
\usepackage{wrapfig}
\usepackage{times}
\usepackage{amssymb}
\usepackage{amsmath}
\usepackage{color}

\usepackage[fleqn]{mathtools}
\usepackage{url}
\usepackage{algorithm}
\usepackage[noend]{algorithmic}
\algsetup{indent=1.3em}

\usepackage{amsmath}
\usepackage{tabularx,booktabs}
\usepackage{array}
\usepackage{graphicx}

\usepackage{threeparttable}
\usepackage{booktabs}
\usepackage{multirow}

\graphicspath{ {./Figures/} }
\newcolumntype{L}{>{$}l<{$}}
\newcolumntype{C}{>{$}c<{$}} 
\newcolumntype{R}{>{$}r<{$}}

\usepackage[font=scriptsize,labelfont=bf, skip=5pt, belowskip=-12pt,position=bottom]{caption}

\newcolumntype{L}{>{\centering\arraybackslash}m{3cm}}

\title{\LARGE\bf Exploring Implicit Human Responses to Robot Mistakes in a Learning from Demonstration Task}

\author{Cory J. Hayes$^{1}$, Maryam Moosaei$^{1}$, and Laurel D. Riek$^{1}$
\thanks{$^{1}$Department of Computer Science \& Engineering, University of Notre Dame, IN, USA
        {\tt\small \{chayes3,mmoosaei,lriek\}@nd.edu}}%
}

\begin{document}

\maketitle

\begin{abstract}
As robots enter human environments, they will be expected to accomplish a tremendous range of tasks. It is not feasible for robot designers to pre-program these behaviors or know them in advance, so one way to address this is through end-user programming, such as learning from demonstration (LfD). While significant work has been done on the mechanics of enabling robot learning from human teachers, one unexplored aspect is enabling mutual feedback between both the human teacher and robot during the learning process, i.e., \textit{implicit learning}. In this paper, we explore one aspect of this mutual understanding, \textit{grounding sequences}, where both a human and robot provide non-verbal feedback to signify their mutual understanding during interaction. We conducted a study where people taught an autonomous humanoid robot a dance, and performed gesture analysis to measure people's responses to the robot during correct and incorrect demonstrations.  
\end{abstract}

\section{Introduction}
\label{Introduction}

Robots are becoming more commonplace in human environments, such as schools, homes, hospitals, and work settings, and  are expected to accomplish a wide variety of tasks. Given the near infinite number of tasks robots might be expected to perform in these varied settings, it is not feasible for robot designers to completely pre-program machines before they are deployed. Many researchers have suggested this problem can be addressed via end-user robot programming, where users can modify and create new behaviors for their robot to best suit their needs and preferences \cite{argall2009,akgun2012}.

Learning from demonstration (LfD) is one such method that enables people to readily develop custom robot behavior \cite{argall2009}. In LfD, a learner automatically creates a mapping between states and actions by watching a teacher perform the task; the learner can then replicate the teacher's actions. The main benefit of LfD is that it is an intuitive way for people to teach robots and does not require the teacher to have highly specialized knowledge, such as the ability to directly program the robot \cite{atkeson1997}. 

There has been significant research in how to design and implement LfD systems, including how people want to teach robots. Work by Thomaz et al. \cite{thomaz2006} showed that LfD systems could be improved  for both the teacher and learner if greater communicative channels could be employed during the learning process. We build upon this work, and specifically are interested in ways to enable human teachers to have more efficient and naturalistic interactions, by way of a common human-human interaction (HHI) phenomena: grounding sequences.

\begin{figure}[t] 
\centerline{\includegraphics[width=.35\textwidth]{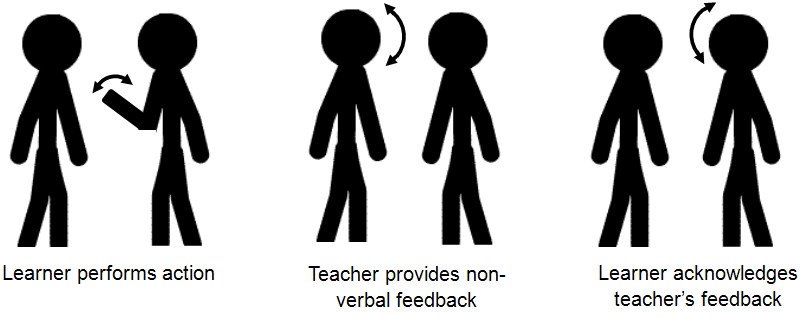}}
\caption{In a grounding sequence, 1) the speaker performs an action, 2) the addressee provides nonverbal backchannel feedback, and 3) the speaker acknowledges this feedback.}
\label{fig:grounding}
\end{figure}

A grounding sequence is a communicative interchange between a speaker and addressee. 
In this exchange, both parties continually provide feedback within the conversation, which enables them to signify whether or not there is a mutual understanding of a topic \cite{clark1991}. Grounding occurs continuously within each moment in conversation, and is not solely confined to pauses in dialogue \cite{bavelas2011}. It is a three-part sequence that occurs when 1) a speaker makes a statement or asks a question, 2) the addressee provides a verbal or nonverbal signal in response to what the speaker has said, and 3) the speaker acknowledges this display (See Fig.~\ref{fig:grounding}). 

During a grounding sequence, the speaker does not simply notice the signal from the addressee and continue talking, but must also acknowledge the signal from the listener by providing an observable behavior in response \cite{bavelas2012}. Grounding is completed when both the speaker and addressee believe that there is a mutual understanding of what has been said.

Clark and Brennan \cite{clark1991} discuss three classes of responses that are used to show positive evidence of grounding. The first type is \textit{acknowledgement}, where back-channel responses, such as a head nod or verbal utterance, is provided by the listener while the speaker is talking. The second type of response is the \textit{relevant next turn} where the speaker gives the listener the chance to directly respond to what has been said, such as asking a question. The third type of response is \textit{continued attention}, where the listener may look away from the speaker and in turn, and the speaker responds by changing his/her dialogue to recapture the attention of the listener. In this paper, we focus on the first type of response, \textit{acknowledgement}.

There have been few studies that have explored the effects of robots generating aspects of grounding sequences in human-robot interaction (HRI). Sidner et al. \cite{sidner2006} performed a study where a conversational robot nodded in response to head nods by participants, and found that people nodded more when the robot performed this action. Krosager et al. \cite{krosager2014} explored the use of nodding as a back-channel response to a human speaker, and found that the physical presence of the robot had a significant impact on user perception when compared to using a virtual agent. Others have explored grounding from the perspective of gaze cues and discourse \cite{billard1998,lemaignan2012,mutlu2009,riekrobinson2008}.

Since the concept of grounding focuses on mutual understanding, it can also be used to facilitate interactions with robots when they are given a task; this can be especially useful to correct robot mistakes that negatively impact users \cite{yasuda2013}. If the user is aware that the robot does not understand a command, the user can adjust the command delivery accordingly \cite{kimleyzberg2009}. Conversely, if a robot is able to detect backchannel feedback from a human, it would be possible for the robot to automatically adjust its behavior, without requiring explicit human feedback. Giving robots the ability to learn from both implicit and explicit feedback may lead more natural and less frustrating interactions by reducing the current complete burden placed on human teachers.

Furthermore, we can uncover principles that would assist in the development of policies to detect, classify, and make robot behavioral decisions based on implicit human feedback.
This point motivates the study described in this paper. Our objective is to observe human behavior during an LfD interaction involving robot mistakes to eventually enable robots to automatically detect when these mistakes occur. 

Intuitively, there is likely a detectable difference in the behaviors people express when robots are performing tasks correctly compared to when they are making mistakes.
The study in this paper focuses on the first step towards creating such a policy: observing what behaviors arise.

In this paper, we explore the occurrences of implicit human feedback in a recorded LfD scenario where a human teaches an autonomous humanoid robot (DARwIn-OP) to perform a dance. The robot detects dance movements from the  human teacher and replicates the movements either correctly or incorrectly. Two independent coders performed a gestural analysis \cite{bavelas2000} of participant's implicit feedback conveyed to the robot, such as via head movements, facial expressions, eye gaze, and body postures. 
Additionally, we gathered qualitative feedback from participants which informed us of ways to further enable human teachers. We discuss these findings in Section~\ref{Results}, and their implication and use for the HRI community in Section~\ref{Discussion}.

\section{Methodology}
\label{Methodology}

We conducted an LfD-centered study to uncover the relationship between a robot's behavior in the first stage of a grounding sequence, and a human's response in the second stage. 
This study is a within-subjects design, where each participant interacts with a robot that performs both correct and incorrect behaviors throughout the interaction.
	
This partial grounding sequence occurs while the robot is demonstrating the moves it has learned from the human teacher. The first stage of the sequence is the robot's nonverbal demonstration of a dance move. The second stage of the sequence is the nonverbal backchannel feedback the human teacher provides while the robot is performing dance moves. We say that our grounding sequence is partial because the third stage where the robot responds to human backchannel feedback does not occur in the current implementation of our study. Instead, our main objective is to help inform the third stage of this grounding sequence.

In the study, participants taught a robot the ``Hokey Pokey'' dance. This is a common dance performed by children in North America, and we chose it for two reasons.  First, due to its repetitive nature, it seemed that it would be easy to learn and recall. We wanted to maximize the user's focus on the robot and less on recalling the mechanics of the task. Second, we wanted to limit the number of true errors participants made, as we employed intentional errors during learning.

\begin{figure*}[t] 
\captionsetup{width=0.8\textwidth}
\centerline{\includegraphics[width=0.8\textwidth]{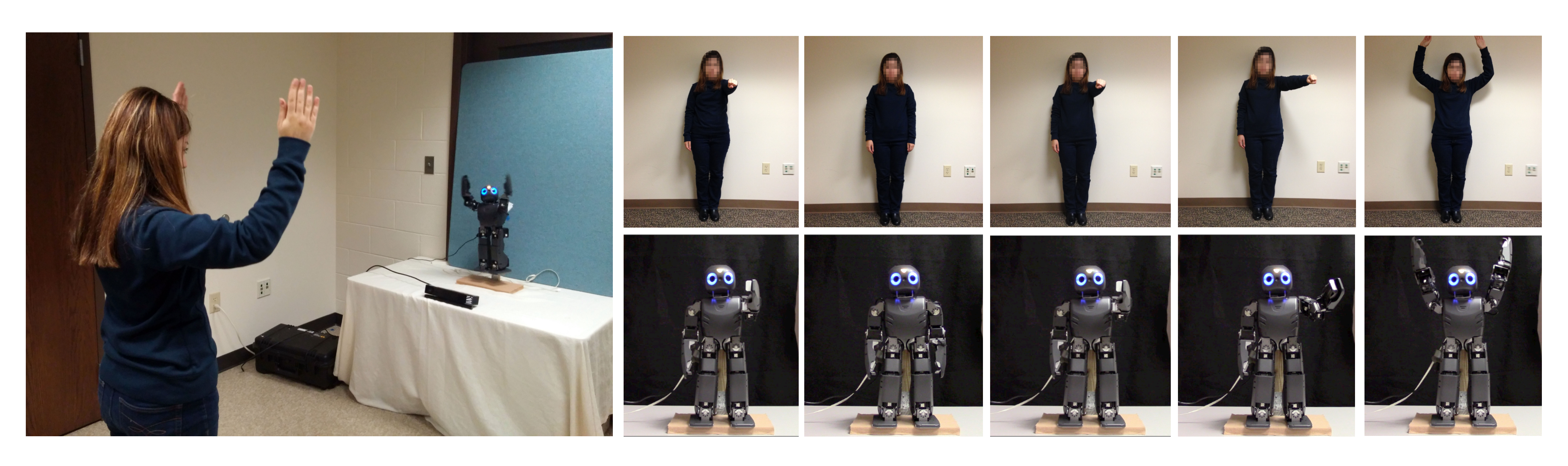}}
\caption{On the left, the setup used for our study showing a person interacting with the robot. On the right, human and robot demonstrations of the hokey pokey for left hand.}
\label{fig:hokeypokey}
\hspace{-0.15in}
\end{figure*}

After considering the motion capabilities of our robot, the final dance consisted of the following sequence: \textit{limb in, limb out, limb in, limb shake, hokey pokey}. For ``limb in'', the participant extends the respective limb towards the robot. For ``limb out'', the person returns to his or her default standing position. ``Limb shake'' is performed by making wide and repeated horizontal movements with the extended limb. Lastly, ``hokey pokey'' consists of the person raising both arms above his or her head and shaking them side to side for a few seconds (See right portion of Fig.~\ref{fig:hokeypokey}). This sequence is conducted across all four limbs (left arm, right arm, left leg, right leg). 

\subsection{Programming and Setup}
\label{Programming and Setup}
Our LfD setup combined the capabilities of a DARwIn-OP humanoid robot and a Microsoft Kinect v2 sensor. 
Using a platform, the robot was positioned in a standing position on a table facing the participant at reasonable human height, as shown in the left side of Fig.~\ref{fig:hokeypokey}.

We created a custom program that automated the interaction through the detection of specific participant movements and basic speech recognition. The Kinect sensor was placed at the base of the robot to limit distraction of its presence, while also being in a position that could reliably collect measurements of participant movements. Based on participant actions detected through 3D point tracking of skeletal joints, the Kinect program compiled robot actions and forwarded commands wirelessly to a custom C++ program running on the robot. 
Participant actions were limited to a set of positions associated with the dance; the program did not respond to any actions outside of this set. 
	
Two RGB cameras were placed in the room to record the interaction. One camera was placed a few feet from the robot, facing the participant with the intention to record facial movements. The second camera was placed behind the participant, facing the robot to supplement the first camera. From this viewpoint, the second camera was able to see both the participant and the robot. 

\subsection{Recruitment}
\label{Recruitment}
We recruited 11 local participants via emails and word-of-mouth, 7 women, 4 men. All participants were native English speakers, and had resided in the United States for an average of 25.64 years (s.d. = 11.16 years). We recruited from this demographic to help ensure prior familiarity with the dance given the way it was implemented in this study. Participants were 27.09 years old on average (s.d. = 9.47).

Since our study focuses on instinctive responses to robot mistakes, we did not inform participants of this true purpose out of concern that this knowledge would influence participant behavior. 
Therefore, the advertised purpose of this study to participants was that we were determining how effectively a person could teach a robot via a LfD task, regardless of their technical background. 

\subsection{Preliminary Tasks}
\label{Prelim}
Prior to the interaction with the robot, each participant completed consent and demographics forms, and were given instructions for the study. To supplement the instruction form, participants also watched a tutorial video depicting an actor teaching the same robot the full dance. In the video, the actor performs the dance and demonstrates what actions to take when the robot performs the dance correctly and when it makes a mistake. Though the video teaches participants how to correct the robot, it does not show the robot making any mistakes or the human actor responding to robot actions beyond demonstrating the dance. At the end of the tutorial video, each participant was instructed to stand facing the robot from a distance of roughly 5-7 feet and the interaction began.


\subsection{Learning from Demonstration Task}
\label{LfDTask}
The LfD dance portion consisted of two stages: training and teaching. We separated the interaction into two stages to give participants the opportunity to learn the dance, as it was implemented for the study, before teaching the robot.

\subsubsection{Training}
\label{Training}
The purpose of the training stage was to allow the participant to practice performing the Hokey Pokey dance moves and have these moves recognized by the Kinect. Participants initiated this stage by raising both arms out from the sides of their bodies in a ``T'' fashion. 
A voiced Kinect program directed participants by stating the move to perform and notified the participant if they did the move correctly or incorrectly. Though the robot is present during this stage, it does not make any signs of activity. At the conclusion of the training stage, the participant is instructed by the Kinect program to once again raise both arms out from their sides to begin teaching the robot.

\subsubsection{Teaching}
\label{Teaching}
Once the participant raises both arms the second time, the DARwIn-OP robot greets the participant by thanking them for their time and stating that it is ready to learn the dance. The Kinect program did not provide any audio output to show activity in this phase similar to how the robot did not make any signs of activity during the training phase. In this stage, the participant teaches the robot the Hokey Pokey dance one movement at a time. After seeing a movement, the robot gives verbal confirmation that it has processed the performed action and that the participant may continue on with the next movement. Participants are informed by both the instruction form and the tutorial video that the ``hokey pokey'' action, which completes the movement sequence for a limb, is a signal for the robot to attempt all of what it has learned so far.

When the robot sees the ``hokey pokey'' action, it announces to the participant that it has detected this movement, will now attempt the dance, and asks the participant to watch it. After performing as much of the dance that it has learned, the robot asks the participant if it did the dance correctly, who in turns responds with a verbal ``affirmative'' or ``negative''. 
If the participant says ``affirmative'', the robot asks which limb it will learn next, in which case the participant sticks out a new limb, waits for the robot's confirmation, returns to their default standing position, waits for another confirmation, and then begins teaching the movements. Similarly, if the participant says ``negative'', the same sequence occurs with the exception being that the presented limb is one that has already been taught. 

As described earlier, our LfD system is designed to repeat any recognized participant movements, regardless of their order, for the purpose of simulating a true LfD scenario. The system is also designed to intentionally make a single apparent mistake during the interaction through a pronounced modification of detected movements. For the intended mistake, the system randomly decides between either adding 3 additional movements for a single limb sequence (e.g. a sequence such as \textit{limb in, out, in, shake, hokey pokey} would become \textit{limb in, out, in, out, in, out, shake, hokey pokey}) or performing just a single movement and immediately going to the ``hokey pokey'' action. The second type of intended mistake can only occur if the limb movement sequence contained at least three actions, not counting the hokey pokey, in order for the truncated move sequence to be noticeable.

From initial testings of the dance interaction, we discovered that we could not reliably pinpoint the exact moments of the interaction where participants observed a robot mistake.
There were instances where there were delayed responses to these mistakes as well as ones where mistakes were completely ignored either willingly to progress or mistakenly due to confusion. Therefore, we set the granularity of identifying mistakes on a per limb basis instead of a per movement basis. If a participant identifies a mistake while the robot performs movements for a specific limb, we consider \textit{all} behavioral responses observed during that limb demonstration to be associated to robot mistakes; the same applies for correct robot behavior.

Since one of our objectives is to explore the behaviors that arise when the robot performs correct and incorrect actions, coders split their behavioral action counts into three intervals for comparisons. The first interval is the ``Correct Interval'', which represents the accumulation of all time intervals per participant where the robot does the correct dance moves for a limb. The second is the ``Incorrect Interval'', which is the accumulation of the time frames where the robot makes a mistake, identified by the participant afterwards. The third is a ``Confirmation Interval'' to represent the times where the robot asks the participant if it has correctly performed the dance, but before the participant gives their verbal response.


\subsection{Post-Interaction}
\label{PostInteraction}
After the robot correctly learned one full iteration of the dance, the interaction ended. Participants were then given an online survey that asked four questions about their perceived teaching abilities during the interaction and the robot's learning abilities. Finally, the participants were given a debriefing form that described the true purpose of the study and a \$5 gift card.

\subsection{Measurement}
\label{Measurements}

Two independent coders employed gestural analysis to label participant interactions using both deductive and inductive coding steps. Deductive coding means that the coders had previous assumptions about the behaviors that would occur in the interaction before conducting the experiment. For example, one would reasonably assume participants would smile, frown, avert their glance, etc. at some point during the interaction. Inductive coding means that coders did not have assumptions prior to conducting the experiment, and created a coding scheme based on observations.

The coders viewed the participant videos and annotated all occurrences of the targeted behaviors during each instance of the robot demonstrating the dance until the participant gives the verbal confirmation to the robot at the end of a dance sequence; these annotations did not include behaviors that occurred while the participant was explicitly instructing the robot as they were trained to do. The coders then categorized the codes (behavior types) into specific hierarchies consisting of gross limb movements, facial movements, self-adaptors, and body postures. They subsequently discussed and resolved any disagreements between codes after analysis, and the recorded data had high inter-rater reliability (\textit{k-alpha = .937}) as calculated on a subset of the data \cite{hayes2007}. 

We focused on the nonverbal, human backchannel feedback and attentiveness behaviors shown in Table \ref{tab:table1} with the corresponding action units from the Facial Action Coding System (FACS) \cite{ekman2005}.

It is worth noting that for this study, coders focused solely on easily observable human responses to robot behavior, and did not attempt to attribute them to any high-level cognitive or emotional states. 
While there has  been previous work done in affective computing and HRI regarding inferring emotions during interaction (c.f., \cite{picard2002,lee2010}, this seemed out of scope and overly restrictive for the current study.

We also did not analyze self-adaptors, which are behavioral responses commonly used to mitigate anxiety, stress, and other emotions \cite{neff2011}.  Examples include scratching, self-grooming, and throat clearing. 

Furthermore, because participants stand in one place throughout the length of our interaction, it is not surprising that frequent body repositioning would happen often, which is more likely to lead to self-adaptive behavior.

\begin{table}[t]

\centering 
\footnotesize
{\begin{tabular}{l l} 
\hline 

      \textbf{Behavior}&\textbf{Description}\\  
\hline\hline 
\textbf{Eyes} &    \\
\multicolumn{1}{m{3cm}}{Glance away from robot (AU61,62)} & \multicolumn{1}{m{3cm}}{Visible eye movement not focused on the robot} \\

Extended eye closures (AU43) & \multicolumn{1}{m{3cm}}{Instances where eyes were closed for at least two seconds} \\ 
\hline 
\textbf{Head} &   \\
Head tilt (AU55,56) & \multicolumn{1}{m{3cm}}{Tilting the top of head towards either shoulder}\\

Raised head (AU53) & \multicolumn{1}{m{3cm}}{Chin lifted, positioned as if looking at some point above}\\

Lowered head (AU54) & \multicolumn{1}{m{3cm}}{Chin lowered, positioned as if looking at some point below}\\

Head turn (AU51,52)& \multicolumn{1}{m{3cm}}{Moving the head to look left/right} \\

Head shake (M60)& \multicolumn{1}{m{3cm}}{Rapid left and right movement of the head}\\

Head nod (M59)& \multicolumn{1}{m{3cm}}{Rapid raised and lowered movement chin movement}\\

\hline 
\textbf{Body} &   \\
Sigh & \multicolumn{1}{m{3cm}}{Shoulders lifted then lowered with visible exhaling motion}\\

Shrug & \multicolumn{1}{m{3cm}}{Shoulders lifted}\\
\hline

\textbf{Mouth} &   \\
Smile (AU12) & \multicolumn{1}{m{3cm}}{Lip corners pulled and raised}\\

\multicolumn{1}{m{3cm}}{Frown(AU 9,10,15,17,20,23,24)}& \multicolumn{1}{m{3cm}}{Lip corners pulled and lowered and/or lips pressed together}\\

Yawn (AU27) & \multicolumn{1}{m{3cm}}{Mouth opened for an extended amount of time}\\
\hline
\textbf{Eyebrows} &   \\
Scrunched Eyebrows (AU4) & \multicolumn{1}{m{3cm}}{Eyebrow(s) lowered, often along with partially closed eyes}\\

Raised Eyebrows (AU1,2) & \multicolumn{1}{m{3cm}}{Eyebrow(s) raised in an arch, often along with widened eyes}\\
[1ex]
\hline 
\end{tabular}}
\caption{Nonverbal behaviors annotated in the experiment and their descriptions.}
\label{tab:table1}
\end{table}


\section{Results}
\label{Results}
Detailed analysis of participants' nonverbal behavior during interaction with the robot revealed several notable features. Averages of the raw data per category are reported in Table~\ref{tab:table3}.

\subsection{General findings} 

\subsubsection{Individual differences in participant expressiveness}

Observations of the recorded videos served as a reminder that individual differences in expressiveness is an important factor to consider when studying human behavior. While most participants displayed a reasonable number of observable behaviors (avg. of 31.91 responses detected per participant) across all three intervals, a few behaved in surprising ways. 

For example, three participants conveyed hardly any of the behaviors in our coding scheme, even when they identified robot mistakes (<12 responses each). They mostly stood still with the same posture and expressions throughout the all robot demonstration instances. It was very difficult to predict their response during the confirmation interval due to the lack of feedback. On the other end, one participant was substantially more expressive than all of the other participants (77.5 responses detected), and it was fairly easy to anticipate whether there would be a ``negative'' or ``affirmative'' confirmation.

\subsubsection{Participant attentiveness}
\label{Attentiveness}

As the interactions progressed, we noticed that some participants paid less attention to the robot after it made at least one mistake. For example, one participant looked away from the robot throughout the entirety of it correctly performing the movements for a specific limb. Two participants retrieved and focused on an item while the robot was demonstrating a portion of the dance, with one participant removing an item from their pocket and the other grabbing the paper instruction form that was left on a desk a few feet behind them.

We suspect this behavior can be attributed to either boredom or frustration with a failing robot; however, as we did not analyze emotions in this study it is not possible to state this with certainty. Informally, three participants verbally mentioned to the researcher that they were frustrated at some point during the interaction.

\subsubsection{Gesture Congruency}
\label{Congruency}

Overall, the behaviors demonstrated by participants were congruous with their verbal confirmations. For example, participants who smiled and nodded along with the robot during a demonstration would typically respond with an ``affirmative'' when the robot asked if it had performed the dance correctly. Similarly, participants who shook their head, frowned, lowered their head with an averted gaze, or closed their eyes for an extended amount of time would typically respond with a ``negative'' for the following robot query.

However, there was one notable example of a participant displaying incongruent behavior that did not match our anticipated response. During one robot demonstration, the participant had a frown that lasted for a few seconds, sighed, closed their eyes for a couple of seconds, glanced away from the robot, and then  frowned again with another glance. 
However, the participant responded with an ``affirmative'' when the robot asked about its correctness.


\subsubsection{Head Nodding}
\label{Nodding}

We also noticed parts of nonverbal grounding sequences for six out of eleven participants, with the primary action being head nods. These sequences rarely occurred at the beginning of the interaction, but became more common as the interaction progressed, especially after a robot made a mistake. For example, one participant displayed an increased focus on the robot, with fewer glances away, while it attempted dance movements on a limb which it had previously made a mistake on. After each movement for this specific limb, the participant nodded their head to acknowledge the correction; however, once the robot made a mistake again, this feedback ended.

\subsection{Questionnaire Results}
\label{Questionnaire}
Participants also completed a post-interaction questionnaire which asked them to reflect on their experiences interacting with the robot. These responses are summarized below. 

\subsubsection{Beliefs about being a good teacher}
\label{Beliefs}
Nearly all participants (9/11) responded affirmatively to the question, \textit{Do you believe you were a good teacher during the interaction with the robot? Why or why not?}. Four participants stated they were good teachers because they believed they were patient with the robot throughout the interaction. Another participant responded similarly, but noted that their patience had waned substantially towards the end of the interaction.

Two participants responded negatively to this question, and stated that the robot did not seem to learn the dance. Four participants partially attributed the robot's failures to their own perceived errors during demonstrations, such as accidentally skipping a dance move or not clearly demonstrating a move to the robot.

\subsubsection{Theory of robot's mind during learning}
\label{Theory}
Participants were asked, \textit{While you were teaching the robot the Hokey Pokey, did you have an idea of how well the robot understood what you were doing and saying? In other words, if the robot could ``think'' like a person does, do you believe you could perceive these thoughts? What did the robot do to make you believe or not believe you could perceive its thoughts?}. 

72\% of participants (8/11) reported that they thought the robot understood the dance movements fairly well until it made its first mistake. Five participants noted that the robot's verbal feedback after each movement demonstration made them believe the robot had a good understanding until mistakes occurred. This verbal feedback simply consisted of utterances such as ``alright'', ``ok'', or ``hmm'' followed by a direct request to continue to the next dance move.

One participant stated, ``No, it didn't feel like he fully understood what I was doing and saying. But then again, I don't think I could perceive its thoughts if it had any. The main thing that made me not believe it was that once I changed my inflection on the word affirmative, and it didn't react to my voice at all.'' 

\subsubsection{Ways the robot could better facilitate teaching}
\label{Facilitation}
Participants were asked, \textit{What other actions could the robot have done to help you be a more effective teacher?}. Participants provided several informative suggestions. First, they suggested it would be helpful to be able to interrupt the robot when it made a mistake while demonstrating the dance. ``It [would be better if it didn't repeat] the whole dance when it was wrong and only the limb that it had a problem with. I would lose focus and get distracted as it repeated the good parts again and then wonder if I had missed a mistake.''

Second, participants requested the robot give more real-time feedback beyond the simple verbal utterances throughout the interaction. They suggested this could help give them better awareness of where they were in the teaching process, and also could reduce frustration. Examples discussed were the robot verbally or visually conveying this information by either stating the limb it had recognized or visually displaying this on a screen. 

Finally, another suggestion was that it would be helpful for the robot to mimic their actions in real time so they could more directly repair a mistake when it occurred. ``Maybe [it could] do the movements along with me so I know that it is understanding as we go along.''

The responses to this question and the previous one address the secondary grounding sequence we focused on to enrich an interaction, where 1) the human teacher demonstrates an action, 2) the robot responds through backchannel feedback, and 3) the teacher acknowledges this feedback.

\subsubsection{Awareness of the true research objective}
\label{Awareness}

As mentioned in Section \ref{Introduction}, the advertised purpose of the study differed from the true purpose due to its nature. The researcher verbally asked each participant whether they realized the study's true intention at any point before being debriefed. None of the participants were aware of this true objective. One participant, with a background in psychology, expressed awareness that there may have been an ulterior motive behind the study, but was not able to determine this motive.


\section{Discussion}
\label{Discussion}

Grounding sequences are an important aspect of face-to-face communication, and might prove invaluable in human-robot interaction. One clear way to incorporate grounding sequences into HRI scenarios is within the space of LfD. Future policies could be created which enable robots to implicitly learn from their human teachers by perceiving their gross motor movements and facial expressions. While this is not always practical from a sensing perspective (occlusion, lighting, etc), it may be straightforward to build systems that can sense simple cues from participants.

The behaviors specified in our coding scheme can serve as a reasonable starting point for robot designers interested in pursuing this path. Given participants' individual differences (and our participant pool), it is not wise to make grand generalizations; however, it does seem that from our data, head nods and smiles appear to commonly be seen during confirmatory teaching sequences, and frowns and head shakes during incorrect ones. 

Glancing away from the robot also seemed to be a meaningful communicative signal during teaching, which aligns with other HRI work \cite{mutlu2009}. However, it can also mean a person is accessing information, or is bored or disengaged. Additional work is needed to understand gaze cues within the context of robot teachers. 
A policy may likely need to consider information contained within combinations of movements and temporal analysis of the interaction itself. For example, a smile alone may be a response to correct robot behavior, but a smile combined with a head tilt and scrunched eyebrows could reflect a response to incorrect robot behavior (possible signal of confusion). Or it is possible that a person may be amused by a robot's mistake the first time it occurs, and therefore displays positive implicit feedback, but reverts to expected negative feedback after a mistake happens one or more additional times. This appeared to happen with some of our participants; so a longitudinal approach may be warranted. 

We noted earlier that there was one prominent instance of a participant providing incongruent feedback which resulted in an unexpected response, as well as participants who provided very little feedback throughout the interaction. These sorts of behaviors will likely be reflected in actual LfD interactions in the future, possibly to a higher degree of ambiguity considering we observed this with just eleven participants. 

Incorporating individual differences may also be vital to give robots the ability to classify implicit human feedback. In addition to differences in expressivity, there may be significant variations in how one's cultural background affects gestures (for example, head nodding / shaking differences between participants from S.E. Asia vs. Europe and the United States). 
A follow-up to this study would incorporate some measures of individual characters, such as a personality assessment, analysis of attitudes towards robots, cultural effects, and so on \cite{salem2014,salem2015}.


As reflected by participant responses, feedback from the robot is vital to provide transparency to users. This has been raised previously in the HRI literature \cite{thomaz2006}, and we too found this in our study. In addition to the robot confirming that it had detected the participant's response, we also added more transparency in the case where a participant gave three consecutive ``negative'' confirmations. When this happened, the robot stated that there must be a mismatch between what it detected and what the participant did, and therefore it would state each individual move it saw until the participant gives an ``affirmative'' confirmation. Participants noted that this kind of feedback was informative and decreased confusion. 


There were a few limitations to our study. First, a number of unintended errors that arose during the interactions that were a result of a combination of human errors and machine recognition errors. We observed a few instances where participants demonstrated multiple moves at once, or individual movements that were ambiguous to the Kinect program. This problem mainly arose from the \textit{limb in} and \textit{limb shake} movements, which are very similar to each other.

To prevent these errors from negatively impacting participants' experiences, we ended each interaction after participants gave a ``negative'' confirmation around 20 minutes into the interaction. While this may have yielded slightly less data, we do not believe this adversely affected our findings. 

In closure, enabling robots to automatically detect implicit human feedback would be a vital ability to allow for more natural interactions with robots and help minimized the communicative burden placed on users. LfD techniques were developed to expand robot usability so that more people can interact with robots, and we seek to facilitate these interactions even further by incorporating implicit human feedback that is automatically generated by users.

The results of this study provided us with valuable information regarding this idea.
We observed examples of positive implicit feedback (smiling, nodding, etc.) being generated as responses to correct robot behavior, and specific types of negative feedback (frowning, averted gazing, etc.) being generated by incorrect robot behavior. However, we also observed behavior that would require careful consideration in future studies, such as noticeable differences in the expressiveness of participants or incongruent behavior that blurs the separation of positive and negative implicit feedback. To further enable human teachers in LfD scenarios, we also gained insight from qualitative responses on how participants desired to teach a robot for this setup, which may be applicable to similar LfD setups. These observations should assist in future studies that focus on implicit human behavior in interactions with robots.



\begin{table}[t]

\centering 
\footnotesize
{\begin{tabular}{l c c c } 
\hline 
      &\multicolumn{1}{m{1.15cm}}{\textbf{Correct Interval}}& \multicolumn{1}{m{1.15cm}}{\textbf{Incorrect Interval}}& \multicolumn{1}{m{1.15cm}}{\textbf{Confirmation Interval}} \\ [0.5ex] 
\hline\hline 
\textbf{Eyes}&   &    &  \\
Glance away from robot  & 0.24 & 0.42 & 0.14 \\
Extended eye closures & 0.03 & 0.05 & 0.01  \\ 
\hline 
\textbf{Head}&   &   &  \\
Head tilt (left/right) & 0.25 &  0.28 &  0.02 \\
Raised head& 0.03 & 0.05 & 0.04\\
Lowered head& 0.06 & 0.07 & 0.03\\
Head turn (left/right) & 0.01 & 0 & 0.01 \\
Head shake& 0.01 & 0.21 & 0\\
Head nod& 0.07 & 0.13 & 0.01\\
\hline 
\textbf{Body}&   &   & \\
Sigh& 0.01 &  0.03 &  0 \\
Shrug& 0 & 0 & 0 \\
\hline 
\textbf{Mouth}&   &  &\\
Smile& 0.07 & 0.15 & 0.01 \\
Frown& 0.21 & 0.53 & 0.09\\
Yawn& 0.03 & 0 & 0 \\
\hline 
\textbf{Eyebrows}&   &   &  \\
Scrunched eyebrows& 0.06 & 0.26 & 0.01\\
Raised eyebrows&  0.01 & 0.05 & 0.03\\
[1ex]
\hline 
\end{tabular}}
\caption{Data was coded into three intervals during the robot demonstration: Correct, Incorrect, and Confirmation. Data reported in this table reflect normalized nonverbal behaviors observed in the study (frequencies divided by the respective numbers of intervals for each interval class). For example, the 0.24 for Glance during the Correct Interval (CI) signifies that \textit{for each individual} CI, there was an average of 0.24 glances.\protect\footnotemark}
\label{tab:table3}
\end{table}

\footnotetext{Note, due to a very small sample size (n = 11), it would be dubious to run statistical means comparisons, and one should not accept a p-value with certainty. Instead, we concur with Gelman \cite{gelman2013commentary} that reliable patterns can be found by averaging, as reported here.}

\section*{Acknowledgment} 

This material is based upon work supported by the National Science Foundation under Grant No. IIS-1253935. The authors also thank Paige Rodeghero.

%
\bibliographystyle{abbrv}
\bibliography{chayesROMAN2016}  
\end{document}